# FEASIBILITY OF GENETIC ALGORITHM FOR TEXTILE DEFECT CLASSIFICATION USING NEURAL NETWORK

Md. Tarek Habib[1], Rahat Hossain Faisal[2], M. Rokonuzzaman[3]

[1]Department of Computer Science and Engineering, Prime University, Dhaka, Bangladesh.
md.tarekhabib@yahoo.com

[2] Department of Electronics and Telecommunication Engineering, Prime University, Dhaka, Bangladesh
rhfaisal@ymail.com

[3]School of Engineering and Computer Science, Independent University, Dhaka, Bangladesh.
zaman.rokon@yahoo.com

## ABSTRACT

*The global market for textile industry is highly competitive nowadays. Quality control in production process in textile industry has been a key factor for retaining existence in such competitive market. Automated textile inspection systems are very useful in this respect, because manual inspection is time consuming and not accurate enough. Hence, automated textile inspection systems have been drawing plenty of attention of the researchers of different countries in order to replace manual inspection. Defect detection and defect classification are the two major problems that are posed by the research of automated textile inspection systems. In this paper, we perform an extensive investigation on the applicability of genetic algorithm (GA) in the context of textile defect classification using neural network (NN). We observe the effect of tuning different network parameters and explain the reasons. We empirically find a suitable NN model in the context of textile defect classification. We compare the performance of this model with that of the classification models implemented by others.*

## 1. INTRODUCTION

The importance of quality control in industrial production is increasing day by day. Textile industry is not an exception in this regard. The accuracy of manual inspection is not enough due to fatigue and tediousness. Moreover, it is time consuming. High quality cannot be maintained with manual inspection. The solution to the problem of manual inspection is automated, i.e. machine-vision-based textile inspection system. Automated textile inspection systems have been drawing a lot of attention of the researchers of many countries for more than a decade. Automated textile inspection systems mainly involve two challenging problems, namely defect detection and defect classification. A lot of research has been done addressing the problem of defect detection, but the amount of research done to solve the classification problem is little and inadequate.

Automated textile inspection systems are real-time applications. So they require real-time computation, which exceeds the capability of traditional computing. Neural networks (NNs) are suitable enough for real-time systems because of their parallel-processing capability. Moreover, NNs have strong capability to handle classification problems. The classification accuracy of an appropriate NN, which handles multiclass problems, is good enough [2, 3]. There is a number of performance metrics of NN models. Classification accuracy, model complexity and training time are three of the most important performance metrics of NN models.

Considering the fact that learning in NNs is an optimization process; genetic algorithm (GA), which is an optimization method, has attracted considerable attention of the NN research community. It has been applied to train NN in many contexts, but, to the best of our knowledge, has not yet been applied in order to classify textile defects.

In this paper, we investigate the feasibility of GA in the context of NN based textile defect classification. We observe and justify the impact of tuning different network parameters, such as crossover rate, mutation rate etc. We attempt to find proper NN model in the context of textile defect classification by tuning these parameters. Finally, we compare the performance of the NN model with that of the classification models described in different published articles in terms of the two performance metrics − accuracy and model complexity.

## 2. LITERATURE REVIEW

A number of attempts have been made for automated textile defect inspection [4-24]. Most of them have concentrated on defect detection, where few of them have concentrated on classification. Mainly three defect-detection techniques [8, 25], namely statistical, spectral and model-based, have been deployed. A number of techniques have been deployed for classification. Among them, NN, support vector machine (SVM), clustering, and statistical inference are notable.

Statistical inference is used for classification in [18] and [19]. Cohen et al. [18] have used statistical test, i.e. likelihood-ratio test for classification. They have implemented binary classification, i.e. categorization of only defective and defect-free. Campbell et al. [19] have used hypothesis testing for classification. They also have implemented classification of only defective and defect-free classes. Binary classification, i.e. categorization of only defective and defect-free fabrics, doesn't serve the purpose of textile-defect classification. Murino et al. [10] have used SVMs for classification. They have worked on spatial domain. They have used the features extracted from gray-scale histogram, shape of defect and co-occurrence matrix. They have implemented SVMs with 1-vs-1 binary decision tree scheme in order to deal with multiclass problem, i.e. distinct categorization of defects. Campbell et al. [17] have used model-based clustering, which is not suitable enough for real-time systems like automated textile inspection systems.

NNs have been deployed as classifiers in a number of articles. Habib and Rokonuzzaman [1] have deployed counterpropagation neural network (CPN) in order to classify four types of defects. They concentrated on feature selection rather than giving attention to the CPN model. They have not performed in-depth investigation on the feasibility of CPN model in the context of automated textile defect inspection.

Backpropagation learning algorithm has been used in [8], [11], [14] and [15]. Saeidi et al. [8] have trained their NN by backpropagation algorithm so as to deal with multiclass problem, i.e. categorizing defects distinctly. They have first performed off-line experiments and then performed on-line implementation. Their work is on frequency domain. Karayiannis et al. [11] have used an NN trained by backpropagation algorithm in order to solve multiclass problem. They have used statistical texture features. Kuo and Lee [14] have used an NN trained by backpropagation algorithm so as to deal with multiclass problem. They have used maximum length, maximum width and gray level of defects as features. Mitropulos et al. [15] have trained their NN by backpropagation algorithm so as to deal with multiclass problem. They have used first and second order statistical features. Islam et al. [9, 23] have used resilient backpropagation algorithm to train their NN. Their networks have been capable of dealing with multiclass problem.

Shady et al. [16] have used learning vector quantization (LVQ) algorithm in order to train their NNs. Their NNs have been implemented in order to handle multiclass problem. They have separately worked on both spatial and frequency domains for defect detection. Kumar [12] has used two NNs separately. The first one has been trained by backpropagation algorithm. The network has been designed for binary classification, i.e. categorization of only defective and defect-free. He has shown that the inspection system with this network was not cost-effective. So he has further used linear NN and trained the network by least mean square error (LMS) algorithm. The inspection system with this NN was cost-effective, but it could not deal with multiclass problem. Inability to deal with multiclass problem doesn't serve the purpose of textile-defect classification. Karras et al. [13] have also separately used two NNs. They have trained one NN by backpropagation algorithm. The other NN used by them was Kohonen's Self-Organizing Feature Maps (SOFM). They have used first and second order statistical-texture features for both NNs. Both of the networks used by them are capable of handling binary classification problem. Categorization of only defective and defect-free fabrics doesn't serve the purpose of textile-defect classification.

## 3. NN MODEL TRAINED GENETIC-ALGORITHM

Learning in NNs can be considered as an optimization process. GA is an optimization method. It can be applied as a learning algorithm on any network topology.

### 3.1. Choice of Activation Function

The GA evaluates the error function at a set of some randomly selected points, which is known as a population, of the definition domain. Taking this information into account, a new set of points, i.e. a new population is generated. Gradually the points in the population approach local minima of the error function. GA can be used when no information is available about the gradient of the error function at the evaluated points. That means the error function does not need to be continuous or differentiable. Therefore, the activation function can also be discontinuous or not differentiable [2].

### 3.2. Initialization of Weights

Initialization of weights is an issue that needs to be resolved. Training begins with initial weight values, which are randomly chosen. Large range of weight values may lead the training phases to take more number of training cycles.

### 3.3. Choice of Fitness Function

An important issue is that how the fitness is measured, i.e. what the definition of fitness function F is. This needs to be resolved. There are many options of defining the fitness function. The goal of classification is to achieve as much accuracy as possible on future, i.e. unseen input or feature vectors [29].

### 3.4. Choice of Convergence Criterion

The stopping or convergence criterion depends on the application although there are a number of options of setting the stopping or convergence criterion of the GA.

### 3.5. Choice of Population Size

The GA evaluates the error function at a set of points in every training cycle. This set of search points is known as the population and the training cycle is known as the generation. The size of the population is represented by M. M has to be chosen as a value less than $2^n$, where n is the number of the bits comprising each search point. A search point is also known as a string in this context and is represented by s [30]. In fact, the proper value of M depends on the application.

### 3.6. Setting Selection Strategy

The GA applies three operators known as genetic operators. Of the three operators, selection is the first operator that comes into play in a training cycle. It determines the strings of current generation, from which the population of next generation is build. The strings selected undergo the crossover operation [30]. The selection operator can be implemented in many ways.

### 3.7. Choice of Crossover Rate

Of the three genetic operators, crossover is the second operator that comes into play in a training cycle. It involves the mixing of two strings. A split point is randomly chosen along the length of either string. The last parts of the two strings are swapped, thereby yielding two new strings [29, 30]. Fig. 1 shows an example of the crossover operation on two 8-bit strings. The split point is 5 here (counting from the left).

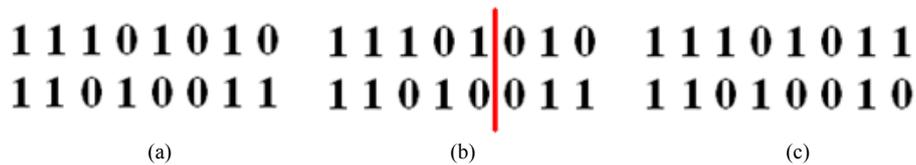

(a)            (b)            (c)

Figure 1  An example of the crossover operation on two 8-bit strings. (a) Two strings are selected. (b) A slit point is randomly chosen. (c) The last parts of the two strings are swapped.

The crossover operator is the most crucial of the three genetic operators in obtaining global result. It is responsible for mixing the partial information contained in the strings of the population [30].

The probability that the crossover operator will be applied on a pair of strings is called the crossover rate $P_c$. If $P_c$ is too low, the average improvement from one generation to the next will be small and the learning will be very long. Conversely, if $P_c$ is too high, the evolution will be undirected and similar to a highly inefficient random search [29]. In fact, the right value of $P_c$ depends on the application. Values between 0.6 and 0.99, inclusive, are reasonable choices of $P_c$ [30].

### 3.8. Choice of Mutation Rate

Mutation is the third and last genetic operator that comes into play in a training cycle. It involves the flipping, i.e. changing from a 1 to a 0 or vice versa, of the bits in a string. Each bit in a string is given a small uniform chance, i.e. probability of being flipped. This small uniform chance is called the mutation rate $P_m$ [29]. Fig. 2 shows an example of the mutation operation on an 8-bit string, where $P_m = 0.01$. A random number, $r \in [0, 1]$, is chosen for each bit of the string 11001001. If $r < P_m$, then the bit is flipped, otherwise no action is taken. For the string 11001001, suppose the random numbers (0.093, 0.041, 0.003, 0.069, 0.027, 0.054, 0.081, 0.009) are generated. Then the bit flips take place. In this case, the third and eighth bits are flipped. The purpose of the mutation operator is to diversify the search and introduce new strings into the population in order to fully explore the search space [30].

Finding the right value of $P_m$ is an important issue that needs to be resolved. If $P_m$ is too low, the average improvement from one generation to the next will be small and the learning will be very long. Conversely, if $P_m$ is too high, the evolution will be undirected and similar to a highly inefficient random search [29]. In fact, the right value of $P_m$ depends on the application. Values between 0.001 and 0.01, inclusive, are reasonable choices of $P_m$ [30].

**Before mutation:** 1 0 ⓪ 0 1 0 0 ①
**After mutation:** 1 0 1 0 1 0 0 0

Figure 2. An example of the mutation operation on an 8-bit string.

### 3.9. Reduction of Computing Units

An important issue is that how large the NN is required to successfully solve the classification problem. This should be resolved. Both training and recall processes take a large amount of time with a large number of computing units. That means computation is too expensive with a large number of computing units. Again, training process does not converge with too small number of computing units. That means the NN will not be powerful enough to solve the classification problem with too small number of computing units [27].

In fact, the right size of NN depends on the specific classification problem that is being solved using NN. One approach to find the right size of NN is to start training and testing with a large NN. Then some computing units and their associated incoming and outgoing edges are eliminated, and the NN is retrained and retested. This procedure continues until the network performance reaches an unacceptable level [27, 28].

## 4. APPROACH AND METHODOLOGY

We are to address the automated textile defect inspection problem. Many possible approaches are investigated in order to accomplish our task. Finally, we have found the approach, shown in Fig. 3, optimal. Our approach starts with an inspection image of knitted fabric, which is converted into a gray-scale image. Then the image is filtered in order to smooth it and remove noises. The gray-scale histogram of the image is formed and two threshold values are calculated from the histogram. Using these threshold values, the image is converted into a binary image. This binary image contains object (defect) if any exists, background (defect-free fabric), and some noises. These noises are removed using thresholding. Then a feature vector is formed calculating a number of features of the defect. This feature vector is inputted to an NN, which is trained earlier by GA with a number of feature vectors, in order to classify the defect. Finally, it is outputted whether the image is defect-free, or defective with the name of the defect.

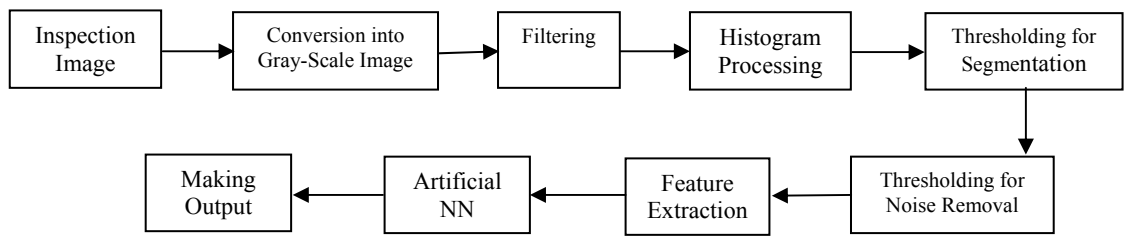

Figure 3. Block diagram of the textile defect inspection method

### 4.1. Defect Types

In this paper, we have dealt with four types of defects. They frequently occur in knitted fabrics in Bangladesh. They are color yarn, hole, missing yarn, and spot shown in Fig. 4. Missing yarn can further be divided into two types – vertical and horizontal [1].

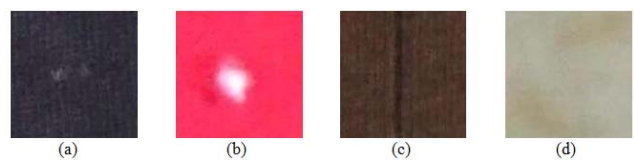

Figure 4. Different types of defect occurred in knitted fabrics. (a) Color yarn. (b) Hole. (c) Missing yarn. (d) Spot.

### 4.2. Terminology

We have adopted some special words [1] for the ease of explanation and interpretation of our automated textile defect inspection problem. We are going to use them in the rest of the paper.

  i) *Inspection Image*: Inspection image or image is the image to be inspected.
  ii) *Defective Region*: Defective Region is the maximum connected area of defect in an inspection image.
  iii) *Defect-Free Region*: Defect-free region is the maximum connected area in an inspection image, which does not contain any defect.
  iv) *Defect Window*: Defect window is the rectangle of minimum area, which encloses all Defective Regions in an inspection image.

### 4.3. An Appropriate Set of Features

An appropriate set of features are selected for classifying the defects. The features are encountered from geometrical point of view. So the features are of same type, namely geometrical feature. Geometrical features describe different discriminatory geometrical characteristics of the defect in the inspection image. The geometrical features selected for classifying the defects are computationally simple to extract. Their discriminatory qualities are also high. Each of these geometrical features is discussed and justified here [1].

  i) *Height of Defect Window, $H_{DW}$.*
  ii) *Width of Defect Window, $W_{DW}$.*
  iii) *Height to Width Ratio of Defect Window, $R_{H/W} = H_{DW} / W_{DW}$* (1)
  iv) *Number of Defective Regions, $N_{DR}$.*

## 5. IMPLEMENTATION

According to our approach to the automated textile defect inspection problem, shown in Fig. 3, we start with an inspection image of knitted fabric of size 512×512 pixels, which is converted into a gray-scale image. In order to smooth the image and remove noises, it is filtered by 7×7 low-pass filter convolution mask, which is shown in Fig. 5. We have tried with a number of masks and find the one in Fig. 4 is the most suitable for our captured images. It considers the pixels in horizontal, vertical and diagonal directions of the center pixel more neighboring than the pixels in all other directions of the center pixels. So, it comparatively accentuates the pixels in horizontal, vertical and diagonal directions of the center pixel. Then gray-scale histogram of the image is formed. From this histogram, two threshold values $\theta_L$ and $\theta_H$ are calculated from the histogram using histogram peak technique [26]. This technique finds the two peaks in the histogram corresponding to the object (defect) and background (defect-free fabric) of the image. It sets one threshold value halfway between the two peaks and the other value either 0 or 255 depending on the positions of the two peaks corresponding to the object (defect) and background (defect-free fabric). Using the two threshold values $\theta_L$ and $\theta_H$, the image with pixels $p(x, y)$ is converted into a binary image with pixels $b(x, y)$, where

$$b(x,y) = \begin{cases} 1, & \text{if } \theta_L \leq p(x,y) \leq \theta_H \\ 0, & \text{otherwise} \end{cases}. \qquad (2)$$

$$\frac{1}{109} * \begin{matrix} 2 & 1 & 1 & 2 & 1 & 1 & 2 \\ 1 & 3 & 2 & 3 & 2 & 3 & 1 \\ 1 & 2 & 4 & 4 & 4 & 2 & 1 \\ 2 & 3 & 4 & 5 & 4 & 3 & 2 \\ 1 & 2 & 4 & 4 & 4 & 2 & 1 \\ 1 & 3 & 2 & 3 & 2 & 3 & 1 \\ 2 & 1 & 1 & 2 & 1 & 1 & 2 \end{matrix}$$

Figure 5. The 7×7 low-pass filter convolution mask

This binary image contains object (defect) if any exists, background (defect-free fabric), and some noises. These noises are smaller than the minimum defect wanted to detect. In our approach, we want to detect a defect of minimum size 3mm×1mm. So, any object smaller than the minimum-defect size in pixels is eliminated from the binary image. If the minimum-defect size in pixels is $\theta_{MD}$ and an object with pixels $o(x, y)$ is of size $S_o$ pixels, then

$$o(x,y) = \begin{cases} 1, & \text{if } S_o \geq \theta_{MD} \\ 0, & \text{otherwise} \end{cases}. \quad (3)$$

Then a number of features of the defect are calculated, which forms the feature vector corresponding to the defect in the image. Fig. 6 shows the images in the corresponding steps mentioned in Fig. 3. Important parts of 512×512-pixel images are shown in Fig. 6 rather than showing the entire images for the sake of space.

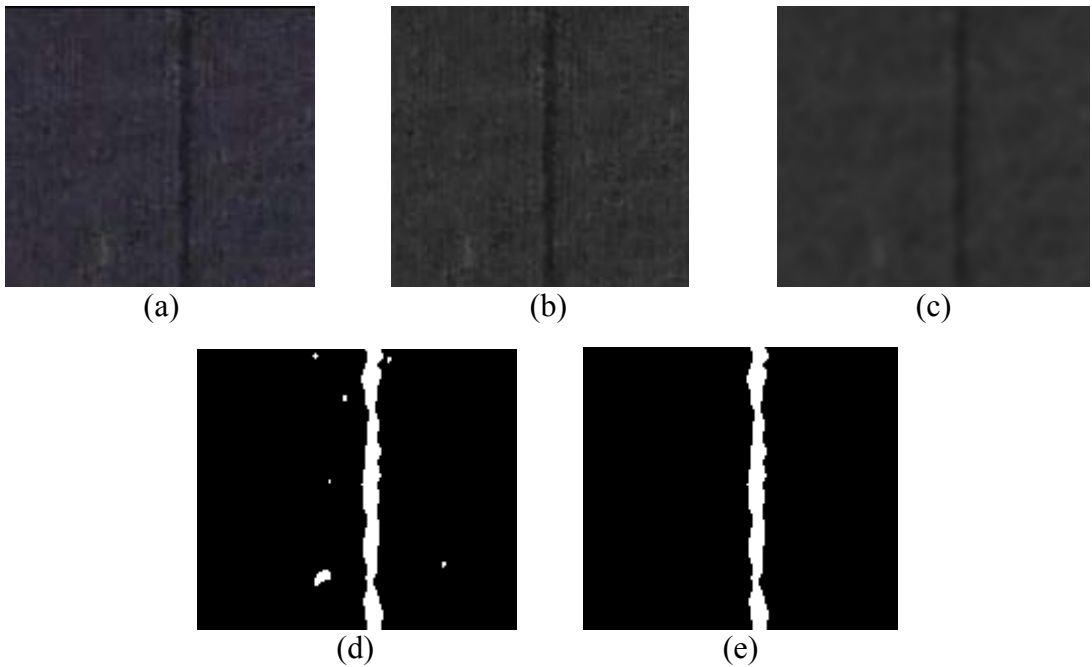

Figure 6. The images of missing yarn in the corresponding steps of our approach. (a) Inspection image. (b) Converted gray-scale image. (c) Filtered image. (d) Segmented image. (e) Noise-removed image.

The classification step consists of the tasks of building a GA model. Building a GA model involves two phases, namely training phase and testing phase. A hundred color images of defective and defect-free knitted fabrics of seven colors are acquired. So, the number of calculated feature or input vectors is 100. That means our sample consists of 100 feature vectors. Table I shows the frequency of each defect and defect-free class in our sample of 100 images.

TABLE I. FREQUENCY OF EACH DEFECT AND DEFECT-FREE CLASS

| No. | Class | Frequency |
|---|---|---|
| 1 | Color Yarn | 6 |
| 2 | Vertical Missing Yarn | 16 |
| 3 | Horizontal Missing Yarn | 16 |
| 4 | Hole | 11 |
| 5 | Spot | 18 |
| 6 | Defect-Free | 33 |
|  | Total | 100 |

The features provided by the feature extractor are of values of different ranges. For example, the maximum value can be 512 for $H_{DW}$ or $W_{DW}$, whereas $N_{DR}$'s can be much less than 512. This causes imbalance among the differences of feature values of defect types and makes the training task difficult for the NN models. According to our context, the scaling, shown in (4), (5), (6), and (7), of the deployed features is made in order to have proper balance among the differences of feature values of defect types. If $H'_{DW}$, $W'_{DW}$, $R'_{H/W}$, and $N'_{DR}$ represent the scaled values of the features provided by the feature extractor, $H_{DW}$, $W_{DW}$, $R_{H/W}$, and $N_{DR}$, respectively, then

$$H'_{DW} = \frac{H_{DW}}{512} \times 100. \tag{4}$$

$$W'_{DW} = \frac{W_{DW}}{512} \times 100. \tag{5}$$

$$R'_{H/W} = 100 \times R_{H/W}. \tag{6}$$

$$N'_{DR} = \sqrt[500]{(N_{DR} - 1) \times 10^{999}}. \tag{7}$$

We split all feature vectors into two parts. One part consisting of 53 feature vectors is for both testing and training the NN model and the other part consisting of the rest of the feature vectors is for testing only. The target values are set to 1 and 0s for the corresponding class and the rest of the classes, respectively. That means if a feature vector is presented to the NN model during training, the corresponding computing unit in the output layer of the corresponding class of the feature vector is assumed to fire 1 and all other units in the output layer are assumed to fire 0. The NN model is trained with maximum number of training cycle $10^6$, maximum amount of training time 5 hours and maximum tolerable error less than $10^{-3}$. That means training continues until $10^6$ training cycles and 5 hours are elapsed and error less than $10^{-3}$ is found. After the training phase is completed, the NN model is tested with all the feature vectors of the both parts. Then all feature vectors are again split into two parts. The first fifty percent of the part for training comes from the previous part for training and the rest fifty percent comes from the previous part for only testing. All other feature vectors form the new part for only testing. The NN model is trained with these new parts and then is tested. In this way, for a specific combination of network parameters, the model is trained and tested 3-5 times in total. We take the results that mostly occur. If the results are uni-modal, we take the results that are the closest to their averages.

We use three-layer feedforward NN for this model, where it is assumed that input layer contributes one layer. We started with a large NN that has 4 computing units in the input layer, 48 computing units in the hidden layer and 6 computing units in the output layer (since we have six different classes as per Table I). We describe the entire training in detail in the following parts of this section, i.e. Section V.

**5.1. Activation Function Chosen**

One of the most used activation functions for GA is the step function,

f : IR → {$x \mid x \varepsilon \{a, b, (a + b) / 2, 0\}$ and $a, b \varepsilon IR$}, which is defined as follows, where $c \varepsilon IR$:

$$f(x) = \begin{cases} a, & \text{if } x < c \\ b, & \text{if } x > c \end{cases} \tag{8}$$

and at c, f(c) is defined to equal a or b or (a + b) / 2 or 0. Common choices are $c = 0$, $a = 0$, $b = 1$, and $c = 0$, $a = -1$, $b = 1$ [27]. In our implementation, we choose the step function, $f : IR \to \{0, 1\}$, which is defined as follows:

$$f(x) = \begin{cases} 0, & if\ x \leq 0 \\ 1, & if\ x > 0 \end{cases}. \qquad (9)$$

### 5.2. Initial Weight Values Chosen

Initialization of weights is an issue that has been resolved. In our implementation, we randomly choose initial weight values of small range, i.e. between -1.0 and 1.0, exclusive, rather than large range, e.g. between -1000 and 1000, exclusive.

### 5.3. Fitness Function Chosen

We want that the goal of classification, which is to achieve as much accuracy as possible on future, i.e. unseen input or feature vectors [29], be reflected in the method of measuring fitness in our implementation. So, we define the fitness function F based on the value of the error function E in the following way:

$$F = \frac{1}{E}. \qquad (10)$$

The value of F will be in $(0, \infty)$ as per (10).

### 5.4. Convergence Criterion Chosen

The stopping or convergence criterion of GA depends on the application. For our implementation of GA, we employ a desired fitness, i.e. inverse of maximum tolerable error, $\theta$, as the convergence criterion. $\theta$ is called the convergence-criterion fitness [29]. We choose the value of $\theta$ as less than $10^{-3}$. That means the training cycle repeats until a search point with fitness greater than $10^3$ is found.

### 5.5. Population Size Chosen

As we mentioned earlier, the proper value of $M$ depends on the application. In our implementation, we first train the NN for $M = n = 64$. We successively increase the value of M and train the NN for that value of M. We find that the fitness and accuracy increase for $64 \leq M \leq 8192$ and start decreasing at M = 10000. We also find that the number of elapsed training cycle increases for $M \geq 64$. So, we choose 8192 as the value of M since we find maximum fitness and accuracy, i.e. 1/7 and 81.44%, respectively, and minimum number of elapsed training cycle, i.e. 952 for this value of $M$.

### 5.6. Selection Strategy Set

As we stated earlier, the selection operator can be implemented in many ways. In our implementation, we focus on allowing the strings with above-average fitness to undergo the crossover operation. That means the average of the fitness of all strings in the population is computed in a training cycle or generation. The strings, which have fitness greater than the average, survive and undergo the operation of crossover [30].

### 5.7. Crossover Rate Chosen

Since values between 0.6 and 0.99, inclusive, are reasonable choices of $P_c$ [30], we first train as well as test the NN for $P_c = 0.99$ and the mutation rate $P_m = 0.01$. We successively decrease the value of $P_c$, and train as well as test the NN for that value of $P_c$, where we keep the value of $P_m$ unchanged. We find that there is no improvement in the fitness and accuracy for $P_c < 0.99$, rather the fitness and accuracy are maximum, i.e. 1/7 and 81.44%, respectively, for $P_c = 0.99$. Although the number of elapsed training cycle is minimum, i.e. 565 for $P_c = 0.9$, we choose 0.99 as the value of $P_c$ because of the accuracy and fitness.

### 5.8. Mutation Rate Chosen

Since values between 0.001 and 0.01, inclusive, are reasonable choices of $P_m$ [30], we first train as well as test the NN for $P_m$ = 0.01 and $P_c$ = 0.99. We gradually decrease the value of $P_m$, and train as well as test the NN for that value of $P_m$ keeping the value of $P_c$ unchanged. We find that there is no improvement in the fitness and accuracy for $P_m$ < 0.01, rather the fitness and accuracy are maximum, i.e. 1/7 and 81.44%, respectively, for $P_m$ = 0.01. Moreover, the number of elapsed training cycle is also minimum for $P_m$ = 0.01. So, we gradually increase the value of $P_m$ from 0.01 and train the NN for that value of $P_m$ keeping the value of $P_c$ unchanged. We find that there is also no improvement in the fitness and accuracy for $P_m$ > 0.01; the fitness and accuracy are maximum, i.e. 1/7 and 81.44%, respectively, for $P_m$ = 0.01. Although the number of elapsed training cycle is minimum, i.e. 803 for $P_m$ = 0.2, we choose 0.01 as the value of $P_m$ because of the accuracy and fitness.

### 5.9. Reduction of Computing Units

As per the approach to find the right size of NN described in Section 3.9, we first train as well as test a large feedforward NN, which has 4 computing units in the input layer, 30 computing units in the hidden layer and 6 computing units in the output layer. Then we successively eliminate 2 computing units in the hidden layer, and train as well as test the reduced NN. We find that there are fluctuations in the fitness function and accuracy as the number of computing units in the hidden layer decreases from 30. The fitness function is the maximum, i.e. 1/5.5 when the number of computing units in the hidden layer is 26 or 14, but the accuracy is the maximum, i.e. 91.75% when the number of computing units in the hidden layer is only 26. We also find that the NNs with 26 and 9 computing units in the hidden layer finish training in minimum and maximum number of cycle, i.e. 50 and 4481, respectively.

## 6. ANALYSIS OF RESULTS

The NN model we implement is for GA. We use three-layer feedforward NN for this model, where it is assumed that input layer contributes one layer. We started with a large feedforward NN, which has 4 computing units in the input layer, 48 computing units in the hidden layer and 6 computing units in the output layer. We describe the results of the entire training, where the number of feature is 4, in detail in the following parts of this section.

### 6.1. Effect of Tuning Population Size

We first train the NN letting the population size ($M$) equal the number of the bits comprising each string in the population, i.e. 64. Then we test the NN with the feature vectors. We successively increase the value of $M$, and train as well as test the NN for that value of $M$. The results achieved are shown in Table II and Fig. 7. Here is to mention that the elapsed time of each training shown in Table II and Fig. 7 is equal to 5 hours.

We see from Table II, Fig. 7(a) and Fig. 7(d) that the fitness function $F$ and accuracy increase for $64 \leq M \leq 8192$ and start decreasing at $M$ = 10000. We also see from Table II, Fig. 7(b) and Fig. 7(c) that the limit of the number of training cycle decreases as $M$ increases, and so is for the number of elapsed training cycle. Larger population size indicates more number of strings, i.e. search points, which means more likelihood of achieving better performance. In our situation, where there are constraints on time, i.e. maximum number of training cycle $10^6$ and maximum amount of training time 5 hours, the fitness function and accuracy get better for $M \leq 8192$ and start getting worse for $M$ > 8192. We also know that larger population size indicates more number of strings, i.e. search points, which means more time for computation in a training cycle. This is why, the limit of the number of training cycle decreases as $M$ increases, and so is for the number of elapsed training cycle.

TABLE II. RESULTS OF TUNING POPULATION SIZE M, WHERE MAXIMUM NUMBER OF TRAINING CYCLE IS $10^6$, MAXIMUM TOLERABLE ERROR IS LESS THAN $10^{-3}$, AND MAXIMUM AMOUNT OF TRAINING TIME IS 5 HOURS

| Network Size | | | Crossover Rate ($P_c$) | Mutation Rate ($P_m$) | Population Size (M) | Fitness Function (F) | Number of Elapsed Training Cycle | Limit of Number of Training Cycle | Accuracy |
|---|---|---|---|---|---|---|---|---|---|
| Input Layer | Hidden Layer | Output Layer | | | | | | | |
| 4 | 30 | 6 | 0.99 | 0.01 | 64 | 1/13.5 | 151461 | 215206 | 60.82% |
| | | | | | 128 | 1/11 | 48526 | 107159 | 67.92% |
| | | | | | 256 | 1/10.5 | 25436 | 53127 | 70.2% |
| | | | | | 512 | 1/9.5 | 19281 | 26547 | 74.23% |
| | | | | | 1024 | 1/8.5 | 2989 | 12855 | 72.16% |
| | | | | | 2048 | 1/7.5 | 2591 | 6625 | 79.38% |
| | | | | | 4096 | 1/7.5 | 1426 | 3233 | 80.41% |
| | | | | | 8192 | 1/7 | 952 | 1584 | 81.44% |
| | | | | | 10000 | 1/9.5 | 110 | 301 | 75.26% |

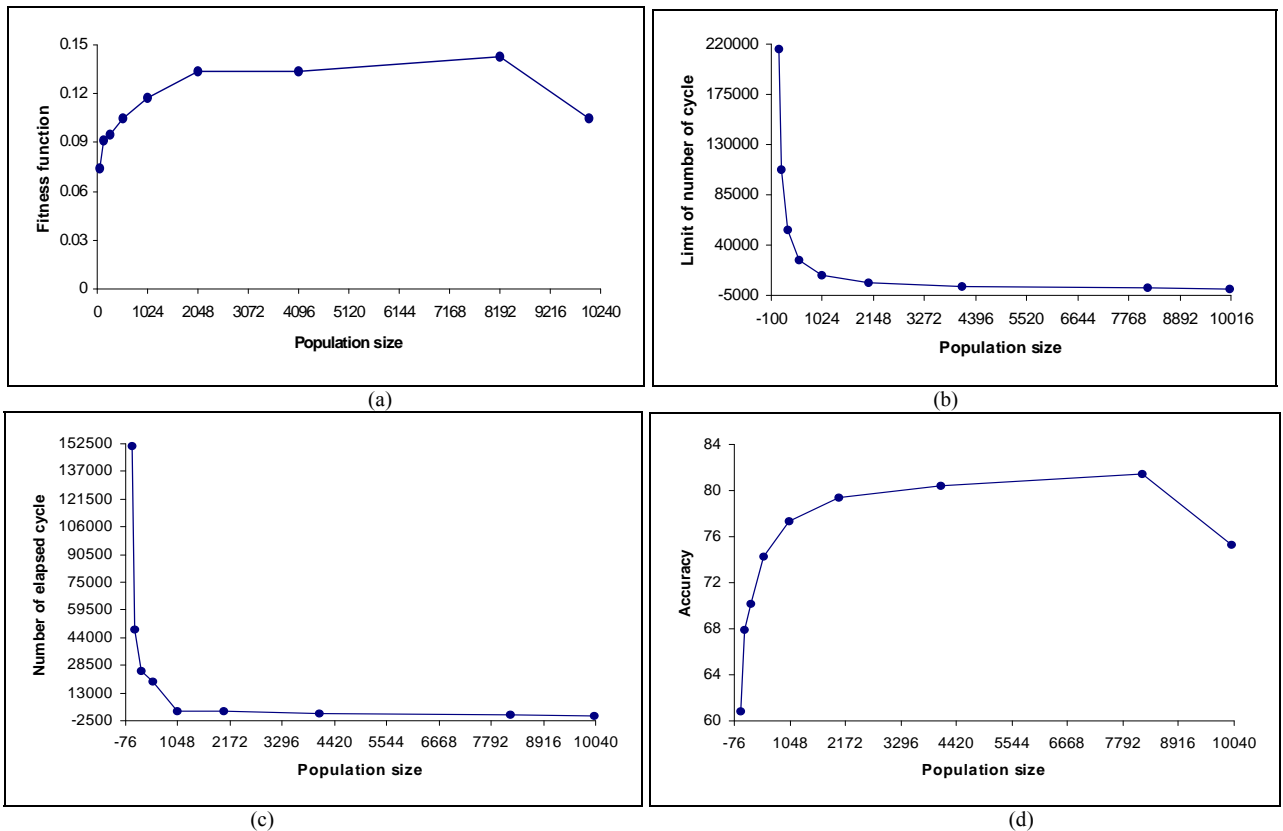

Figure 7. Results of tuning population size M, where maximum number of training cycle is $10^6$, maximum tolerable error is less than $10^{-3}$, and maximum amount of training time is 5 hours. (a) Fitness function F. (b) Limit of number of elapsed training cycle. (c) Number of elapsed training cycle. (d) Accuracy.

## 6.2. Effect of Tuning Crossover Rate

We first train the NN letting the crossover rate ($P_c$) equal 0.99 and the mutation rate ($P_m$) equal 0.01. Then we test the NN with the feature vectors. We successively decrease the value of $P_c$, and train as well as test the NN for that value of $P_c$ keeping the value of $P_m$ unchanged. The results achieved are shown in Table III and Fig. 8. Here is to mention that the elapsed time of each training shown in Table III and Fig. 8 is 5 hours.

We see from Table III, Fig. 8(a) and Fig. 8(d) that there are fluctuations in the fitness function F and accuracy for $0.5 \leq P_c \leq 0.99$, and they are the maximum, i.e. 1/7 and 81.44%, respectively, for $P_c = 0.99$. We also see from Table III and Fig. 8(b) that the limit of the number of training cycle increases as $P_c$ decreases from 0.99, but there are fluctuations in the number of elapsed training cycle as $P_c$ decreases from 0.99, and the number of elapsed training cycle is the minimum, i.e. 565 for $P_c = 0.9$.

Smaller value of $P_c$ indicates less probability of performing the crossover operation, which means less time for computation in a training cycle. This is why, the limit of the number of training cycle increases as $P_c$ decreases from 0.99. We also know that values between 0.6 and 0.99, inclusive, are reasonable choices of $P_c$ [30]. In this context, $F$ and the accuracy are the maximum, i.e. 1/7 and 81.44%, respectively, for $P_c = 0.99$.

TABLE III. RESULTS OF TUNING CROSSOVER RATE $P_c$, WHERE MUTATION RATE $P_m$ IS 0.01, MAXIMUM NUMBER OF TRAINING CYCLE IS $10^6$, MAXIMUM TOLERABLE ERROR IS LESS THAN $10^{-3}$, AND MAXIMUM AMOUNT OF TRAINING TIME IS 5 HOURS

| Network Size (No. of Computing Units) | | | Crossover Rate ($P_c$) | Mutation Rate ($P_m$) | Population Size (M) | Fitness Function (F) | Number of Elapsed Training Cycle | Limit of Number of Training Cycle | Accuracy |
|---|---|---|---|---|---|---|---|---|---|
| Input Layer | Hidden Layer | Output Layer | | | | | | | |
| 4 | 30 | 6 | 0.99 | 0.01 | 8192 | 1/7 | 952 | 1585 | 81.44% |
| | | | 0.9 | | | 1/9 | 565 | 1595 | 76.29% |
| | | | 0.8 | | | 1/9.5 | 626 | 1602 | 72.16% |
| | | | 0.7 | | | 1/8.5 | 1267 | 1607 | 77.32% |
| | | | 0.6 | | | 1/9 | 1490 | 1611 | 75.26% |
| | | | 0.5 | | | 1/9.5 | 598 | 1615 | 72.16% |

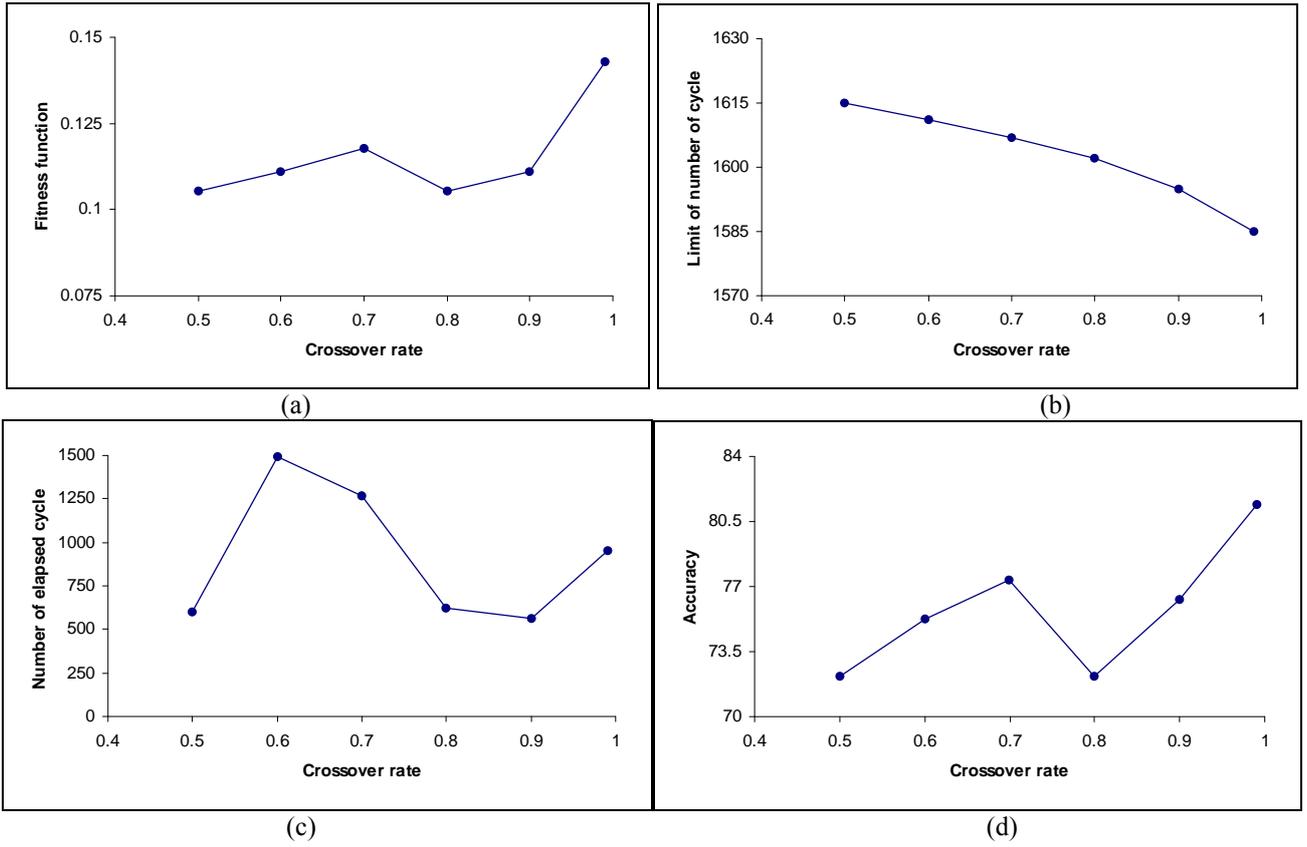

Figure 8. Results of tuning crossover rate $P_c$, where $P_m$ is 0.01, maximum number of training cycle is $10^6$, maximum tolerable error is less than $10^{-3}$, and maximum amount of training time is 5 hours. (a) Fitness function F. (b) Limit of number of elapsed training cycle. (c) Number of elapsed training cycle. (d) Accuracy.

### 6.3. Effect of Tuning Mutation Rate

We first train the NN letting the crossover rate ($P_m$) equal 0.01 and the mutation rate ($P_c$) equal 0.99. Then we test the NN with the feature vectors, which comprise our entire sample. We gradually decrease the value of $P_m$, and train as well as test the NN for that value of $P_m$ keeping the value of $P_c$ unchanged. The results achieved are shown in Table IV and Fig. 9. Again, we gradually increase the value of $P_m$ from 0.01, and train as well as test the NN for that value of $P_m$ keeping the value of $P_c$ unchanged so that improved fitness function and accuracy can be found. The results achieved are shown in Table V and Fig. 10. Here is to mention that the elapsed time of each training shown in Table IV, Fig. 9, Table V and Fig. 10 is 5 hours.

We see from Table IV, Fig. 9(a) and Fig. 9(d) that there are fluctuations in the fitness function $F$ and accuracy for $0.001 \leq P_m \leq 0.01$, and they are the maximum, i.e. 1/7 and 81.44%, respectively, for $P_m = 0.01$. We also see from Table IV and Fig. 9(b) that the limit of the number of training cycle increases as $P_m$ decreases from 0.01, but there are fluctuations in the number of elapsed training cycle as $P_m$ decreases from 0.01, and the number of elapsed training cycle is the minimum, i.e. 952 for $P_m = 0.01$. Again, we see from Table V, Fig. 10(a) and Fig. 10(d) that there are fluctuations in $F$ and accuracy for $0.01 \leq P_m \leq 0.5$, and they are the maximum, i.e. 1/7 and 81.44%, respectively, for $P_m = 0.01$. We also see from Table V, Fig. 10(b) and Fig. 10(c) that the limit of the number of training cycle decreases as $P_m$ increases from 0.01, but there are fluctuations in the number of elapsed training cycle as $P_m$ increases from 0.01, and the number of elapsed training cycle is the minimum, i.e. 803 for $P_m = 0.2$.

Smaller value of $P_m$ indicates less probability of performing the mutation operation, which means less time for computation in a training cycle. This is why, the limit of the number of training cycle increases as $P_m$ decreases from 0.01. Conversely, the limit of the number of training cycle decreases as $P_m$ increases from 0.01. We also know that values between 0.001 and 0.01, inclusive, are reasonable choices of $P_m$ [30]. In this context, F and the accuracy are the maximum, i.e. 1/7 and 81.44%, respectively, for $P_m$ = 0.99.

TABLE IV. RESULTS OF TUNING MUTATION RATE $P_m$ BELOW 0.01 (INCLUSIVE), WHERE CROSSOVER RATE $P_c$ IS 0.99, MAXIMUM NUMBER OF TRAINING CYCLE IS $10^6$, MAXIMUM TOLERABLE ERROR IS LESS THAN $10^{-3}$, AND MAXIMUM AMOUNT OF TRAINING TIME IS 5 HOURS

| Network Size (No. of Computing Units) | | | Crossover Rate ($P_c$) | Mutation Rate ($P_m$) | Population Size (M) | Fitness Function (F) | Number of Elapsed Training Cycle | Limit of Number of Training Cycle | Accuracy |
|---|---|---|---|---|---|---|---|---|---|
| Input Laye | Hidden Layer | Output Layer | | | | | | | |
| 4 | 30 | 6 | 0.99 | 0.001 | 8192 | 1/9 | 1536 | 1597 | 75.26% |
| | | | | 0.0025 | | 1/8 | 1038 | 1596 | 79.38% |
| | | | | 0.004 | | 1/9 | 1462 | 1595 | 75.26% |
| | | | | 0.0055 | | 1/7.5 | 1004 | 1593 | 80.41% |
| | | | | 0.007 | | 1/9.5 | 1443 | 1591 | 73.19% |
| | | | | 0.0085 | | 1/8.5 | 954 | 1588 | 78.35% |
| | | | | 0.01 | | 1/7 | 952 | 1585 | 81.44% |

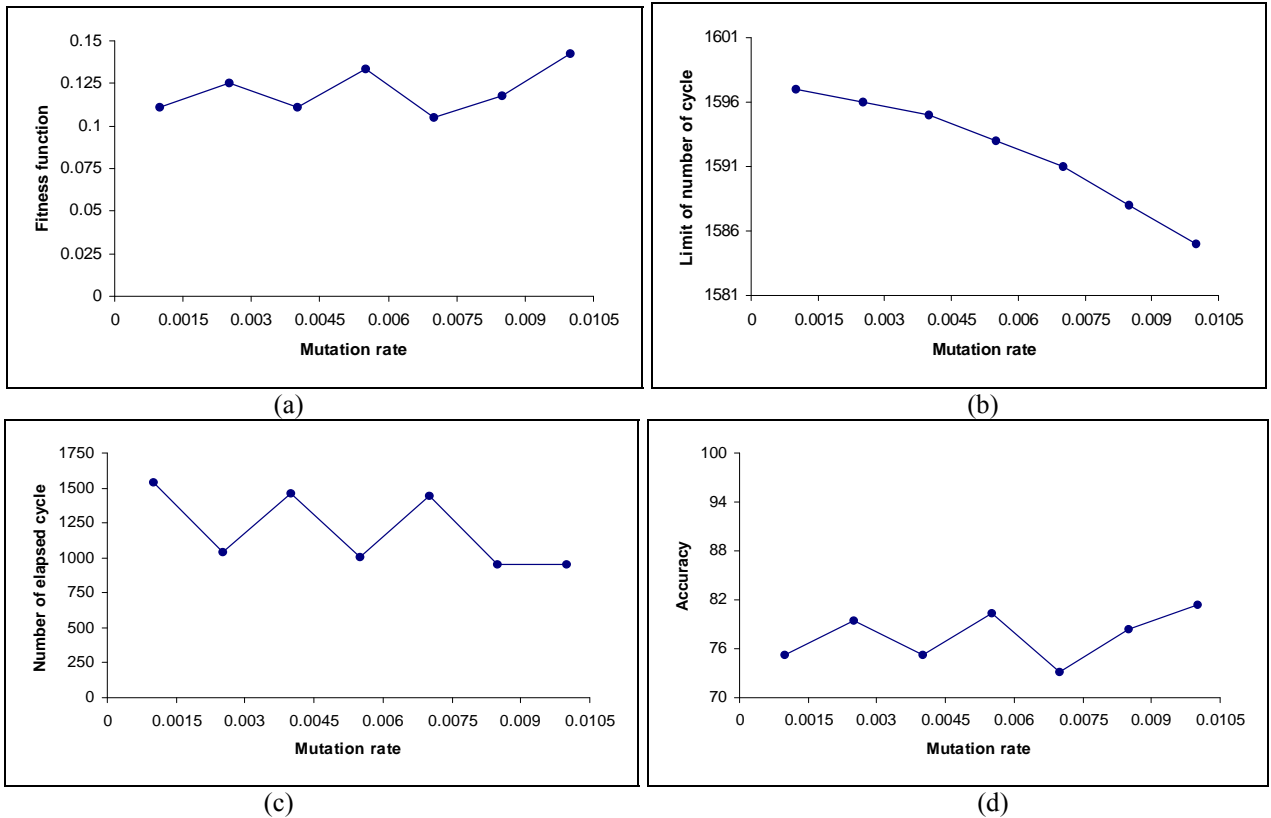

Figure 9. Results of tuning mutation rate $P_m$ below 0.01 (inclusive), where $P_c$ is 0.99, maximum number of training cycle is $10^6$, maximum tolerable error is less than $10^{-3}$, and maximum amount of training time is 5 hours. (a) Fitness function F. (b) Limit of number of elapsed training cycle. (c) Number of elapsed training cycle. (d) Accuracy.

TABLE V. RESULTS OF TUNING MUTATION RATE $P_m$ ABOVE 0.01 (INCLUSIVE), WHERE CROSSOVER RATE $P_c$ IS 0.99, MAXIMUM NUMBER OF TRAINING CYCLE IS $10^6$, MAXIMUM TOLERABLE ERROR IS LESS THAN $10^{-3}$, AND MAXIMUM AMOUNT OF TRAINING TIME IS 5 HOURS

| Network Size (No. of Computing Units) | | | Crossover Rate ($P_c$) | Mutation Rate ($P_m$) | Population Size (M) | Fitness Function (F) | Number of Elapsed Training Cycle | Limit of Number of Training Cycle | Accuracy |
|---|---|---|---|---|---|---|---|---|---|
| Input Layer | Hidden Layer | Output Layer | | | | | | | |
| 4 | 30 | 6 | 0.99 | 0.01 | 8192 | 1/7 | 952 | 1585 | 81.44% |
| | | | | 0.1 | | 1/9.5 | 926 | 1560 | 73.19% |
| | | | | 0.2 | | 1/8 | 803 | 1515 | 79.38% |
| | | | | 0.3 | | 1/8.5 | 895 | 1492 | 77.32% |
| | | | | 0.4 | | 1/7.5 | 1194 | 1475 | 80.41% |
| | | | | 0.5 | | 1/8 | 970 | 1459 | 78.35% |

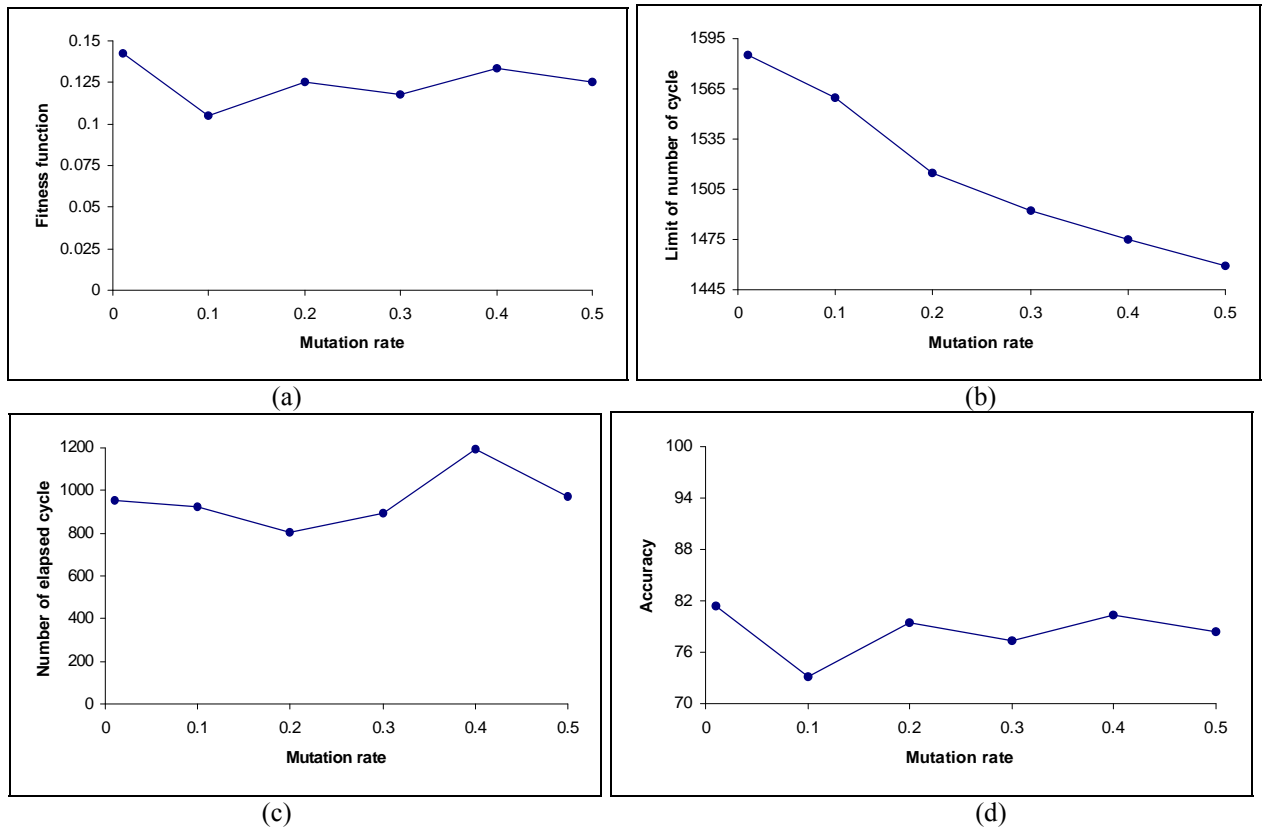

Figure 10. Results of tuning mutation rate $P_m$ above 0.01 (inclusive), where $P_c$ is 0.99, maximum number of training cycle is $10^6$, maximum tolerable error is less than $10^{-3}$, and maximum amount of training time is 5 hours. (a) Fitness function F. (b) Limit of number of elapsed training cycle. (c) Number of elapsed training cycle. (d) Accuracy.

## 6.4. Effect of Reducing Computing Units

We first train a large feedforward NN, which has 4 computing units in the input layer, 30 computing units in the hidden layer and 6 computing units in the output layer, and we test the NN with the feature vectors, which comprise our entire sample. Then we successively eliminate 2 computing units in the hidden layer, and train as well as test the reduced NN. We carry on the procedure until the network performance reaches an unacceptable level. The results achieved are shown in Table VI and Fig. 11. Here is to mention that the elapsed time of each training shown in Table VI and Fig. 11 is 5 hours.

We see from Table VI, Fig. 11(a) and Fig. 11(d) that there are fluctuations in the fitness function F and accuracy as the number of computing units in the hidden layer decreases from 30. F is the maximum, i.e. 1/5.5 when the number of computing units in the hidden layer is 26 or 14, but the accuracy is the maximum, i.e. 91.75% when the number of computing units in the hidden layer is only 26. We also see from Table VI and Fig. 11(b) that the limit of the number of training cycle increases as the number of computing units in the hidden layer decreases from 30, but there are fluctuations in the number of elapsed training cycle as the number of computing units in the hidden layer decreases from 30, as shown in Table VI and Fig. 11(c). We see from Table VI and Fig. 11(c) that the NNs with 26 and 9 computing units in the hidden layer finish training in minimum and maximum number of cycle, i.e. 50 and 4481, respectively.

TABLE VI. RESULTS OF REDUCING COMPUTING UNITS IN HIDDEN LAYER, WHERE CROSSOVER RATE $P_c$ IS 0.99, MUTATION RATE $P_m$ IS 0.01, MAXIMUM NUMBER OF TRAINING CYCLE IS $10^6$, MAXIMUM TOLERABLE ERROR IS LESS THAN $10^{-3}$, AND MAXIMUM AMOUNT OF TRAINING TIME IS 5 HOURS

| Network Size (No. of Computing Units) | | | Crossover Rate ($P_c$) | Mutation Rate ($P_m$) | Population Size (M) | Fitness Function (F) | Number of Elapsed Training Cycle | Limit of Number of Training Cycle | Accuracy |
|---|---|---|---|---|---|---|---|---|---|
| Input Layer | Hidden Layer | Output Layer | | | | | | | |
| 4 | 30 | 6 | 0.99 | 0.01 | 8192 | 1/7 | 952 | 1585 | 81.44% |
| | 28 | | | | | 1/9 | 710 | 1740 | 75.26% |
| | 26 | | | | | 1/5.5 | 50 | 1826 | 91.75% |
| | 24 | | | | | 1/7 | 1295 | 1991 | 80.41% |
| | 22 | | | | | 1/10.5 | 169 | 2187 | 68.04% |
| | 20 | | | | | 1/9 | 1070 | 2405 | 76.29% |
| | 18 | | | | | 1/7.5 | 200 | 2666 | 79.38% |
| | 16 | | | | | 1/8 | 2007 | 2999 | 78.35% |
| | 14 | | | | | 1/5.5 | 349 | 3412 | 89.69% |
| | 12 | | | | | 1/8 | 732 | 4020 | 79.38% |
| | 10 | | | | | 1/8 | 4077 | 4781 | 78.35% |
| | 9 | | | | | 1/7 | 4481 | 5311 | 80.41% |
| | 8 | | | | | 1/9 | 1464 | 5943 | 76.29% |

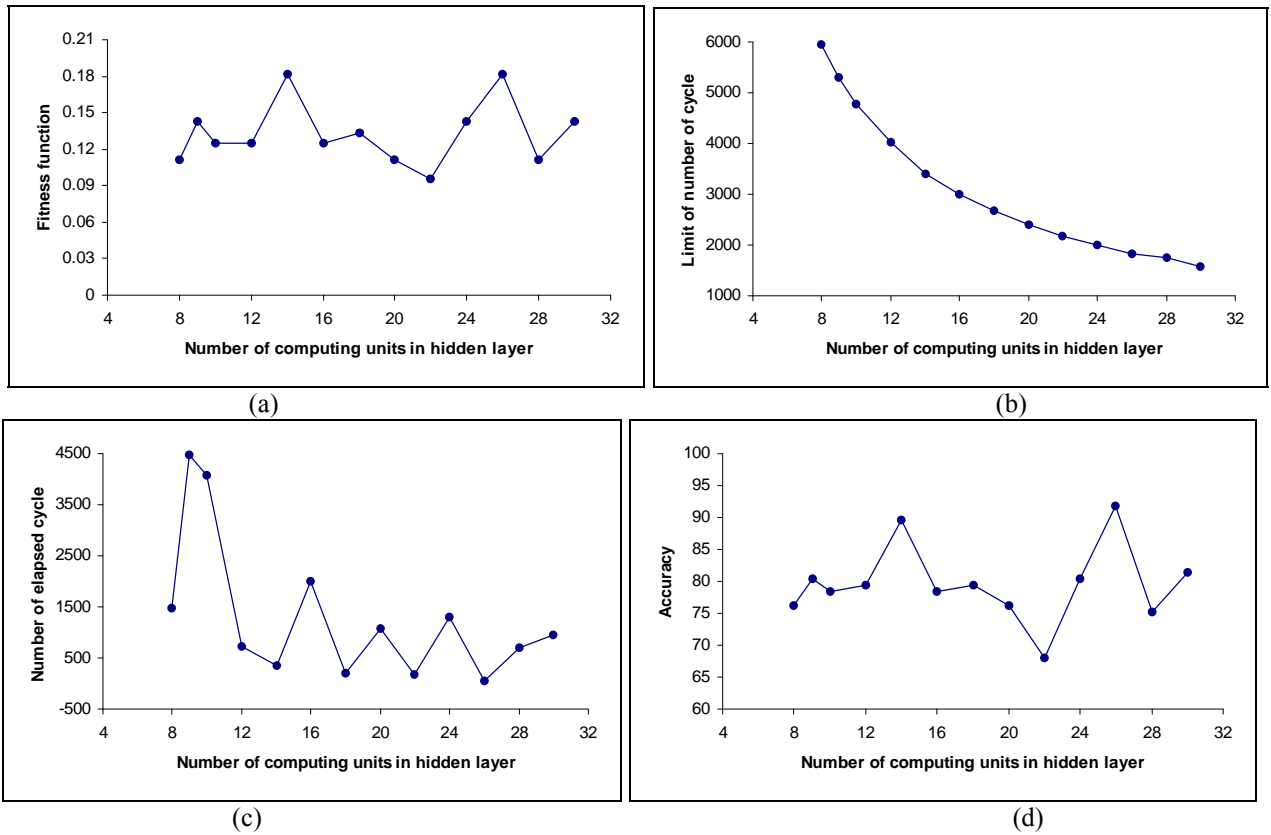

Figure 11. Results of reducing computing units in hidden layer, where $P_c$ is 0.99, $P_m$ is 0.01, maximum number of training cycle is $10^6$, maximum tolerable error is less than $10^{-3}$, and maximum amount of training time is 5 hours. (a) Fitness function F. (b) Limit of number of elapsed training cycle. (c) Number of elapsed training cycle. (d) Accuracy.

Considering two of the most important performance metrics of NN models, namely accuracy and model complexity, we come up with the decision that the NN that contains 26 computing units in the hidden layer is the best in our context. In case of this NN, the accuracy is modest (91.75%) and the model complexity, i.e. the number of computing units is not small (4-26-6).

## 7. COMPARATIVE ANALYSIS OF PERFORMANCE

We need to compare our GA-trained NN model with others' models in order to have a proper understanding of our model. The models implemented by others are for the environment and constraints that may not be same as or similar to ours. It is difficult to compare the models others implemented with ours. Hence, substantially comparative comments cannot be made. Nevertheless, we try to perform comparison as much substantial as possible.

SVMs have been used by Murino et al. [10] for classification. Two data sets, i.e. sets of images have been separately used in their entire work. One set contained 2 types of fabric and the other set contained 4 types of fabric. In neither case, name of any fabric type is mentioned. However, size of data set in both cases was adequate. The first set contained 1117 images, where the second one contained 1333 images. They have got 99.11% and 92.87% accuracy for the first and second set of images, respectively. Although good accuracy has been achieved for the first set of images, the accuracy achieved for the second one is modest.

NNs have been deployed as classifiers in a number of articles, where none has performed a detailed investigation of the feasibility of NNs they used in the context of textile defect classification. Habib and Rokonuzzaman [1] have trained their CPN in order to classify defects commonly occurring in knitted fabrics. They used four types of defects and two types of features. Their sample consisted of 100 images. Their CPN had 4, 12 and 6 computing units in the input, hidden and output layers respectively. It took 191 cycles for the CPN to be trained. A 100%-accuracy has been found. Although the accuracy and model complexity (number of computing units) have been good and medium respectively, the training time has been long.

Backpropagation learning algorithm has been used in [8], [11], [14] and [15]. Saeidi et al. [8] have worked with knitted fabrics. They have first performed off-line experiments and then performed on-line implementation. In case of off-line experiments, the sample size was 140. They have employed a three-layer feedforward NN, which had 15, 8 and 7 computing units in the input, hidden and output layers respectively. It took 7350 epochs for the NN to be trained. An accuracy of 78.4% has been achieved. The model complexity (number of computing units) has been modest. Moreover, the training time has been long and the accuracy has been poor. In case of on-line implementation, the sample size was 8485. An accuracy of 96.57% has been achieved by employing a feedforward NN. The accuracy has been good although the model complexity and training time have not been mentioned. Karayiannis et al. [11] have worked with web textile fabrics. They have used a three-layer NN, which had 13, 5 and 8 computing units in the input, hidden and output layers respectively. A sample of size 400 was used. A 94%-accuracy has been achieved. Although the accuracy and model complexity have been good and small respectively, nothing has been mentioned about the training time. Kuo and Lee [14] have used plain white fabrics and have got accuracy varying from 95% to 100%. The accuracy has been modest. Moreover, the model complexity and training time have not been reported. Mitropulos et al. [15] have used web textile fabrics for their work. They have used a three-layer NN, which had 4, 5 and 8 computing units in the input, hidden and output layers respectively. They have got an accuracy of 91%, where the sample size was 400. The accuracy has been modest although the model complexity has been small. Nothing has been mentioned about the training time. Resilient backpropagation learning algorithm has been used in [9] and [23]. Islam et al. [9] have used a fully connected four-layer NN, which contained 3, 40, 4, and 4 computing units in the input, first hidden, second hidden and output layers respectively. They have worked with a sample of over 200 images. They have got an accuracy of 77%. The accuracy has been poor and the model complexity has been large. Moreover, the training time has not been given. Islam et al. [23] have employed a fully connected three-layer NN, which had 3, 44 and 4 computing units in the input, hidden and output layers, respectively. 220 images have been used as sample. An accuracy of 76.5% has been achieved. The accuracy and model complexity have been poor and large respectively. Moreover, nothing has been mentioned about the training time. Shady et al. [16] have separately worked on both spatial and frequency domains in order to extract features from images of knitted fabric. They have used the LVQ algorithm in order to train the NNs for both domains. A sample of 205 images was used. In case of spatial domain, they employed a two-layer NN, which contained 7 computing units in the input layer and same number of units in the output layer. They achieved a 90.21%-accuracy. The accuracy has been modest although the model complexity has been small. Moreover, the training time has not been reported. In case of frequency domain, they employed a two-layer NN, which had 6 and 7 computing units in the input and output layers respectively. An accuracy of 91.9% has been achieved. Although the model complexity has been small, the accuracy has been modest. Moreover, nothing has been mentioned about the training time.

With respect to such observation, our obtained accuracy of more than 91% and model complexity of (4-26-6) appear to be promising enough. If maximum amount of training time were much more than 5 hours, much good accuracy and model complexity could have been found. As we have mentioned before, due to the lack of uniformity in the image data set, performance evaluation and the nature of intended application, it is not prudent to explicitly compare merits of our approach with other works. Therefore, it may not be unfair to claim that GA has enough potential to classify textile defects with very encouraging accuracies.

## 8. CONCLUSION AND FUTURE WORK

In this paper, we have investigated the feasibility of GA-trained NN model in the context of textile defect classification. We have observed and justified the impact of tuning different network parameters. We have attempted to find proper GA model in the context of textile defect classification by tuning these parameters. Finally, we have compared the performance of the GA model with that of the classification models described in different articles in terms of the performance metrics - accuracy and model complexity.

Due to small sample size, our finding is not comprehensive enough to make conclusive comment about the merits of our implemented GA model. There remains work with GA to successfully classify commonly occurring all types of textile defects for a sample of a very large number of high quality images.